# Belief Rule Based Expert System to Identify the Crime Zones

Abhijit Pathak[1], Abrar Hossain Tasin[2]

*Abstract*— **This paper focuses on Crime zone Identification. Then, it clarifies how we conducted the Belief Rule Base algorithm to produce interesting frequent patterns for crime hotspots. The paper also shows how we used an expert system to forecast potential types of crime. In order to further analyze the crime datasets, the paper introduces an analysis study by combining our findings of the Chittagong crime dataset with demographic information to capture factors that could affect neighborhood safety. The results of this solution could be used to raise awareness of the dangerous locations and to help agencies predict future crimes at a specific location in a given time.**

## I. INTRODUCTION

Crimes are a common social issue which affects a society's quality of life and economic growth. It is considered an important factor deciding how people migrate to a new city and what areas should be avoided while traveling. When crime increases, law enforcement agencies continue to demand innovative expert solutions to enhance crime detection and protect their communities better. While crimes may occur anywhere, it is normal for offenders to focus on the crime opportunities they face in the most familiar places. Through offering a Belief Rule Expert System to identify the crime zones and to evaluate the type of the location and time of the crimes committed, we hope to raise awareness of the dangerous locations in certain periods of time [1].

Therefore, our proposed solution, along with saving lives, can potentially help people stay away from the locations at a certain time of day. Therefore, gaining this kind of awareness will help people make better choices about their places of living. Police forces, on the other hand, can use this approach to improve crime detection and prevention. In addition, this would be useful for allocating policing resources. It can help to ensure an efficient use of police resources in the distribution of police at most likely places of crime for any given time. By providing all of this information, we hope to make our community safer for the people who live there as well as for others who will travel there.

The expected results cannot be assured of 100 percent accuracy but the results show that our program helps to some degree to minimize the crime rate by providing security in sensitive areas of crime. So we have to collect and review crime reports to create such an effective crime analytics tool [2].

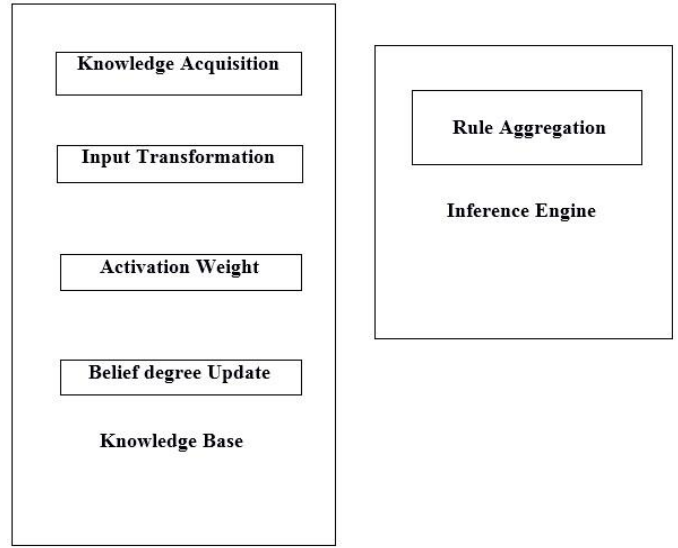

Fig 1: A Typical Belief Rule Based Expert (BRB) System

Belief Law The expert system is an evolution of the standard law-based system and is capable of representing more complex casual relationships using various types of information with ambiguity. A degree of belief is associated with every possible consequence of a rule in an expert system based on belief system rules. In the creed rule based expert system, the expert system based knowledge is developed using evidential reasoning approach which is the replacement of the knowledge base inferential engine. Complicated and continuous casual relationships between different factors can be captured by the creed rule base expert system that traditional IF-THEN rules cannot apply [3].

Research would include evaluating, developing, implementing, and reviewing the best system for defining Crime environment. It will concentrate on the processes performed to define crime zones for many reasons, and how to determine how to detect crime hotspots before decisions are made on which and when hotspots can be dangerous. It will also focus on how normal people and the law enforcement can know the crime hotspots properly.

Actually, the study of the crime zone and the detection of the effects of the crime zone is based on paper. Our study is aimed at finding spatial and temporal criminal hotspots using a set of real-world crime datasets. We must try to locate the most likely places for the crime and their frequent occurrence. We will also predict what type of crime could occur next in a specific location within a given time frame [4].

Develop a computer-based system that best identifies the crime zone and analyzes the results of crime hotspot tests and the underlying conditions to recommend the best safe zone for the people.

- Characterize and significance of hotspot selection process.
- Develop an expert user-friendly system with specific knowledge to advise a person on the selection of crime zone identifiers.
- Assess the performance of the Expert System developed.

In order to achieve the expected goals, the following objectives were addressed:

- Defining system requirements to assess the information needs of different users so that the best crime hotspot identification system can be delivered effectively and effectively.

- Assess the processes involved in the monitoring of the crime zone and the parameters used to assess the crime areas can be unsafe for ordinary people.

- Develop a new concept for the crime zone identification system that analyzes the hotspot and not only provides specific feedback on crime hotspots but also measures possible crime zone.

## II. RELATED WORK

Countless work relating to crime has been done. Large datasets were reviewed and information such as location and type of crime was extracted to help people follow the law enforcement. Existing methods used those databases to identify location-based crime hotspots. There are several map applications showing the exact location of the crime along with the type of crime for any given city. Although crime locations have been identified, there is no information available that includes the date and time of the crime, as well as techniques that can predict exactly what crimes will occur in the future [5].

On the other hand, the previous work and its existing methods mainly identify crime hotspots based on the location of high crime density without taking into account either the crime type or the date and time of the crime. Related research work, for example, containing a dataset for the city of Philadelphia with crime information from 1991-1999. It focused on the existence of complex multi-scale relationships between time and space [6]. Another study titled "The effectiveness of hotspot analysis to forecast spatial crime trends" looks at the various types of crime to see if they vary in their predictive capabilities. Many studies are investigating the relationship between criminal activity and socio-economic variables such as age, race, income and unemployment [7].

None of them consider the three elements (location, time, type of crime) together, in spite of all the existing work. In our study we provide a Belief Rule Based Expert System to Identify the Crime Zones.

## III. METHODOLOGY

In proposed system various components are interact to each other. Their proper interaction creates perfects BRB and ER based system. During implementation of BRB and ER methodology first important things is to choose programming language to implement ER methodology and user interface, knowledge based need to implement using database. After designing perfect Graphical User Interface (GUI), it will be easy to general user to interact with the developed system and perform meditation assessment [8]. Thus, the implementation of the proposed system of experts is one of the main tasks to be tackled for this work. It is understood that BRB and ER methods for the system proposed can be applied using an expert knowledge-based system. Expert knowledge-based system works with data from input and output. Interaction between the different components of the knowledge-based expert system for our proposed system is shown in the following:

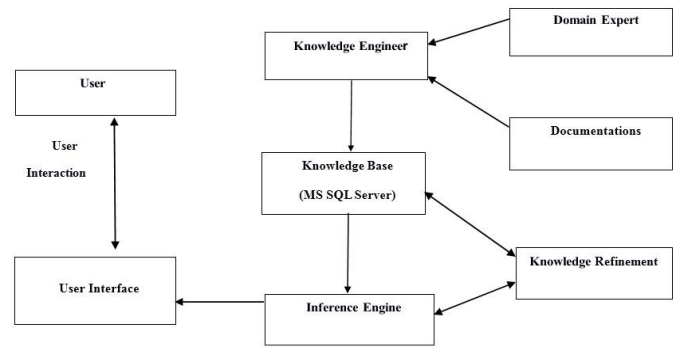

Fig 2: System Architecture

It's known that BRB and ER methodology for the proposed system is possible to implement using a knowledge based expert system. Knowledge based expert system works with input and output data. Interaction among various components of knowledge based expert system for our proposed system is shown in below:

- The crime zone identifier system must have a database that can support the storage and retrieval of user details, crime details and details that users who log in to the system will access later.
- The crime zone identification system must have a client interface that allows privileged users, such as administrators, to perform tasks such as adding / editing criminal information, adding / editing hotspot details and creating / editing user details.

- The crime zone identifier system must have a viewing functionality to allow normal and privileged users to view the details of a particular entity from the system database.
- The crime zone identification program must also have a software interface for normal users to sign up for user accounts, enter data for review and measurement.
- The identifying crime zone system must have a functionality that produces summary reports from analyses and calculations.

*A. Crime Zone Prediction*

Crime zone prediction is a systematic approach for assessment and analyzing patterns and trends in crime. This system can predict places which have high probability for crime occurrence and can also reveal crime prone areas. With the increasing advent of computerized systems, identifying of crime zone can help the Law enforcement officers to speed up the process of solving crimes [8].

*B. Crime Zone*

Crime zone means the areas which have high crime intensity. Although crimes could occur everywhere, it is common that criminals work on crime opportunities they face in most familiar areas for them. We hope to raise people's awareness about the dangerous locations in certain time periods.

*C. Purpose of identifying Crime Zone*

Identification of crime zone can help the Law enforcement officers to speed up the process of solving crimes. By using Belief Rule Based Expert System we can extract previously unknown, useful information from an unstructured data.

*D. Types of Crime Zone*

There are four types of Crime zone—
- Places
- Victims
- Streets
- Areas

The most basic form of a hot spot is a place that has many crimes. A place can be an address, street corner, store, house, or any other small location, most of which can be seen by a person standing at its center. Crime zone prediction is a systematic approach for assessment and analyzing patterns and trends in crime. This system can predict places which have high probability for crime occurrence and can also reveal crime prone areas. With the increasing advent of computerized systems, identifying of crime zone can help the Law enforcement officers to speed up the process of solving crimes [9].

*E. Crime Zone Prediction Techniques*

There are many techniques for predicting crime zone. From all techniques three most popular techniques for identifying crime zone given below:
- Maps and Geographic Information Systems
- Statistical Tests
- Reading a Crime Map

*F. Factors related to Crime Zone*

- Type of offence;
- Date and time of the incident;
- Location of the incident; (like outside visitor rate, dark alleys, resident density etc.)
- Type of premises where the incident occurred;
- Whether drugs or alcohol were involved;
- Whether a weapon was used;
- The age and gender of the offender;
- The age and gender of the victim, and more. (like women population)
- Traffic in the area.
- Income level in the area. (Unemployment, population density)
- Education rate in the area.
- Social cause (Poverty, inequality etc.)

*G. Knowledge Acquisition by BRB*

Belief rule base (BRB) is the extension of traditional rule base, because BRB added an additional belief structure. The belief structure knowledge represents parameters including rule weights, attribute weight and belief degree in consequent are embedded. The difference between BRB and Rule base in belief makes BRB more flexible. Although a traditional IF-THEN rule and a belief IF-THEN rule are same. The main difference is in accuracy. In traditional If-THEN rule the consequent is either 100% true or 100% false. But in belief rue base, it's capable to represent causal relationship among antecedence and consequent, so rate of accuracy enhances in belief rule base. On the other hand, traditional If-THEN rule is not capable to handling this kind of relationship between antecedence and consequents, so we can say that such kind of relationship can be represent in a better way by using belief structure. Here lies the importance of using BRB over traditional IF-THEN rule [10].

*H. Design of BRB for Identifying Crime Zone*

Knowledge base creation is an important part of Identifying Crime Zone. Knowledge acquired from several sources to create the knowledge base. Those sources includes domain experts and written documentations. Knowledge base creation involving several steps to deal with, as- first of all BRB framework need to develop, according to which the work will go further. After the BRB development, initial rule base creation will be followed. Then the assessment process will move forward with data transformation process, which will involve input transformation, activation weight and belief degree update process [11].

To design the BRB for expert system, first analyze various factors involved into crime zone prediction and effect of those factor. In this step, came to learn about factors like land type, water removal situation, drainage system, soil texture and pH value etc. Then discuss with domain exporters to validate proposed BRB framework and after validation finally got the following framework for identifying crime zone.

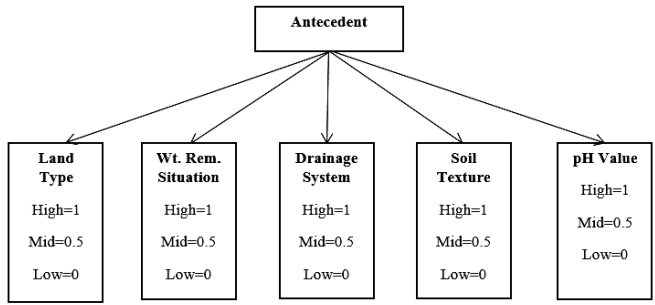

Fig 4: Antecedent for Identifying Crime Zone

It has been shown how to calculate input transformation in Chapter-3. Input transformation process enables the system to transform human readable input data into system readable data for best crop selection. It's been possible due to the use of RIMER methodology in proposed system, which is a contribution of this project in best crop selection process. For instance, let's take the consequent X (behavioral impact), for which one of antecedent Land Type, Water Removal Situation, Drainage System, Soil Texture and pH value Evolution grade for each factor are:

$$0.7 <= High <= 1.0$$
$$0.4 <= Mid <= 0.6$$
$$0.0 <= Low <= 0.3$$

Let take the following portion of BRB as example to show initial rule base construction:

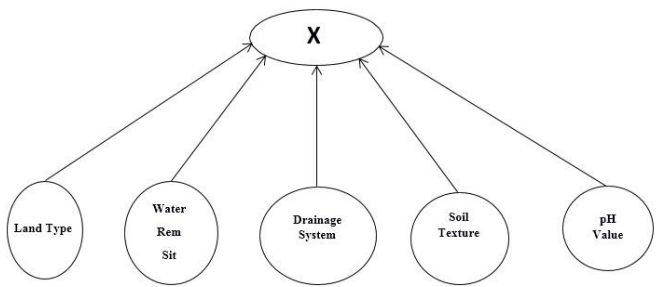

Fig 5: Input Data

## IV. IMPLEMENTATION OF EXPERT SYSTEM

In proposed system various components are interact to each other. Their proper interaction creates a perfects BRB and ER based system. During implementation of BRB and ER methodology first important things is to choose programming language to implement ER methodology and user interface, knowledge based need to implement using database. After designing perfect Graphical User Interface (GUI), it will be easy to general user to interact with the developed system and perform meditation assessment. So implementation of proposed expert system is one of the major task to deal with for this work. System implementation was achieved using MySQL, JAVA scripting language. The database was designed using MySQL because it is highly efficient and an effective solution for data storage and handling even in a networked environment [12].

### A. Initial Rule Base for X

X is behavioral impact, it's the consequent which has five antecedents are Land Type, Water Removal Situation, Drainage System, Soil Texture and pH value. Land Type has 3 referential values (high, mid and low). Water Removal Situation has 3 referential value (early, avg. and late). Drainage System has 3 referential values (well, good and poor). Soil Texture has 3 referential values (sandy, silt and clay). pH value has 3 referential value (acidic, neutral and alkynes).

Table 1: Evaluation Grade for Behavioral Impact

| Types of Land | Wt. Removal Situation | Drainage System | Soil Texture | pH value |
|---|---|---|---|---|
| 1 = High | 1 = Early | 1 = Well | 1 = Sandy | 1 = Acid |
| 0.5= Mid | 0.5 = Average | 0.5 = Good | 0.5 = Silt | 0.5=Neutral |
| 0 = Low | 0 = Late | 0 = Poor | 0 = Clay | 0= Alkynes |

### B. User Interface

It already been said that, proposed system developed using JAVA and MY SQL Server, eclipse is complete GUI driven IDE, so developer don't need to worry about efficiency of interface or data handling, eclipse IDE offers important GUI components need to use to develop system; button, data grid, form, checkbox, combo list, label etc. so most of the system development done in visual designing way with little back-end coding [13]. Following is a snapshot of developed system.

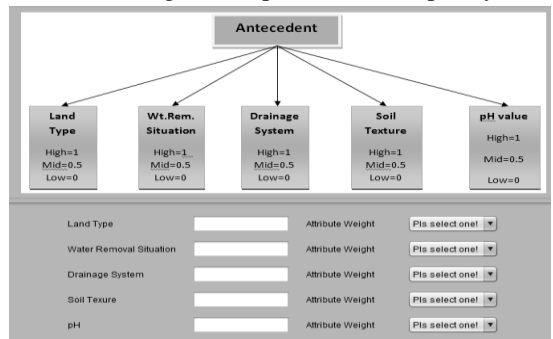

Fig 6: User Interface

Each node (button) works as a link to a certain form for data input, without calculating nodes in subsequent order, user can't move further, as it's been seen that some nodes are grayed out. Syntax grid showing the syntax for corresponding nodes. After calculating all nodes, aggregated values for corresponding input value will be shown in nodes. Finally aggregated fuzzy values will be converted into numerical value to get the desired result, which will indicate a value in scale of meditation assessment [14].

## V. RESULT AND DISCUSSION

Data for Identifying Crime Zone of the BRB framework in (previous) have been collected from the CMP. Fifty expert data have been collected and for simplicity some of them have been shown in Table. The expert has selected the major five major factor of nine factor and the associated risk factors data from the expert. The collected data of the expert has been used as input data in this expert system to assess the crop selection. The BRBES has been compared with a fuzzy rule based expert system (FRBES), developed in the MATLAB environment. Illustrates the results of crime zone prediction carried out by the BRBES (Column 2), the expert (Column 3), and by the FRBES (Column 4). If the value of expert is greater or equal 50 positive than benchmark considered as "1" and otherwise it is considered as "0". This data has been considered as the benchmark data as shown in column 5 of the table.

Table 2: Crime Zone Assessment by BRBES, EXPERT and RBFL System

| SL No. | BRBES | EXPERT | RBFL | BENCHMARK |
|---|---|---|---|---|
| 1 | 79.95 | 69.44 | 74.34 | 1 |
| 2 | 89.39 | 70.32 | 78.23 | 1 |
| 3 | 73.15 | 60.13 | 67.15 | 1 |
| 4 | 58.1 | 45.15 | 46.23 | 1 |
| 5 | 40.45 | 30.45 | 22.25 | 0 |
| 6 | 44.52 | 37.57 | 38.54 | 0 |
| 7 | 55.01 | 50.65 | 51.55 | 1 |
| 8 | 41.57 | 45.56 | 42.26 | 0 |
| 9 | 33.27 | 30.29 | 30.25 | 0 |
| … | …….. | …….. | …….. | …….. |
| … | …….. | …….. | …….. | …….. |
| 48 | 91.85 | 79.24 | 82.37 | 1 |
| 49 | 61.54 | 53.53 | 56.24 | 1 |
| 50 | 24.15 | 18.73 | 21.45 | 0 |

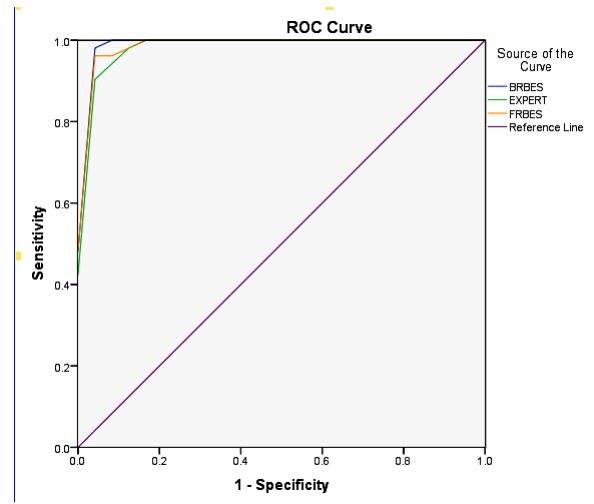

Fig 7: Comparison of Reliability among BRBES, FRBES and Expert Opinion by using ROC curves

The Receiver Operating Characteristics (ROC) curve is widely used to analyze the effectiveness of assessment having ordinal or continuous results. Therefore, it has been considered in this research to test the accuracy of the BRBES output against expert opinion and FRBES. The accuracy or performance of the BRBES in crop selection can be measured by calculating the Area under Curve (AUC). If AUC of BRBES output is larger than the expert opinion and FRBES then it can be inferred that BRBES produce more accurate and reliable results. Fig. 7 illustrates three ROC curves, one represents the performance of the BRBES, one represents the performances of the expert and the other represents the performance of FRBES. The ROC curve plotted by the blue line in Fig. 8.1 represents the result generated by the BRBES and its area under curve (AUC) is 0.853(95% confidence intervals 0.708 - 0.999) while the curve with green line represents the results generated by the expert opinion and its area under curve (AUC) is 0.7604(95% confidence intervals 0.721 - 0.967) and the curve with grey line represents the results generated by the FRBES and its area under curve (AUC) is 0.803(95% confidence intervals 0.709-0.988). From the AUC of the BRBES', expert opinion and FRBES it can be observed that AUC of BRBES is greater than that of FRBES as well as from expert opinion.

Table 3: Area under the Curve

| Test Result Variable(s) | Area |
|---|---|
| BRBES | .998 |
| EXPERT | .990 |
| FRBES | .995 |

The reason for having less AUC against expert opinion in Crime Zone Prediction is that expert is not aware of the uncertainty associated with the factor and risk factors of Crime Zone selection. Moreover, during our research I have observed while asking expert about the procedures of measuring factor, we have noticed there approach is Boolean. The reason for less performance of the Fuzzy rule based expert system than from the belief rule based expert system is that belief rule base considers in addition to uncertainty and incompleteness. In addition, the inference procedures of BRBES consists of input transformation, rule activation weight calculation, belief update and rule aggregation using evidential reasoning approach [15]. Evidential reasoning is capable of process various types of uncertainties, which is not the case with the fuzzy based inference engine.

In this paper, we have demonstrated the design, development and application of BRBES to identify crime zone by taking five major factors of crime zone prediction. This BRBES employed a novel methodology known as RIMER, allows the handling of various types of uncertainties. Hence, the results generated from the BRBES is reliable than from expert opinion or from fuzzy rule based expert system [16].

## VI. CONCLUSION

The project was successful in implementing the objectives stipulated in earlier chapters. This system offers a number of benefits to the users; normal persons and researchers can automatically register, update and view records whereas the system administrator can manage and configure the various parameters of system functionality. In addition, the system can also authenticate the users, display the status of a crop yields and total costs and generate reports. In order to improve and increase the use of Information Technology Solutions in day to day life , there is a need for further research in order to:-

- Come up with more advanced software that can remotely connect all the other remote major towns in districts round and globally.
- Enable a normal person or researcher to use the internet, log in to this system and remotely analyze test results and recommend crime hotspots and get all the information they need say, total danger and so on.
- Other demands may arise with the ongoing evolution and use of the system.

The study demonstrated how a short text specification of a system can be modeled in analysis, expanded and detailed into a design model, and finally implemented and programmed in PHP, HTML and JavaScript. The various parts of this study were designed by group members who made every effort to work in the same manner they would have done on an actual business project. And though the different phases and activities might seem separate and; to have been conducted in a strict sequence, the work is more iterative in practice.